\pdfoutput=1

\documentclass[11pt]{article}

\usepackage[final]{acl}

\usepackage{times}
\usepackage{latexsym}
\usepackage{multirow}
\usepackage{booktabs}
\usepackage{amsmath,amsfonts}
\usepackage{algorithmic}
\usepackage{algorithm}
\usepackage{graphicx}
\usepackage{bm}
\usepackage{subfigure}
\usepackage{float}

\usepackage[T1]{fontenc}

\usepackage[utf8]{inputenc}

\usepackage{microtype}

\usepackage{inconsolata}

\title{Enhancing Reinforcement Learning with Label-Sensitive Reward for Natural Language Understanding}

\author{Kuo Liao\footnotemark[1],\quad Shuang Li\footnotemark[1],\quad Meng Zhao,\quad Liqun Liu\footnotemark[2], \\ {\bf Mengge Xue,} \quad {\bf Zhenyu Hu,}\quad {\bf Honglin Han,}\quad {\bf Chengguo Yin} \\
Tencent\\
\texttt{\{magialiao,shuangsali,liqunliu\}@tencent.com}}

\begin{document}
\maketitle
\begin{abstract}
Recent strides in large language models (LLMs) have yielded remarkable performance, leveraging reinforcement learning from human feedback (RLHF) to significantly enhance generation and alignment capabilities. However, RLHF encounters numerous challenges, including the \textit{objective mismatch} issue, leading to suboptimal performance in Natural Language Understanding (NLU) tasks.
To address this limitation, we propose a novel Reinforcement Learning framework enhanced with Label-sensitive Reward (RLLR) to amplify the performance of LLMs in NLU tasks. By incorporating label-sensitive pairs into reinforcement learning, our method aims to adeptly capture nuanced label-sensitive semantic features during RL, thereby enhancing natural language understanding.
Experiments conducted on five diverse foundation models across eight tasks showcase promising results. In comparison to Supervised Fine-tuning models (SFT), RLLR demonstrates an average performance improvement of 1.54\%. Compared with RLHF models, the improvement averages at 0.69\%. These results reveal the effectiveness of our method for LLMs in NLU tasks. Code and data available at: \href{https://github.com/MagiaSN/ACL2024_RLLR}{https://github.com/MagiaSN/ACL2024\_RLLR}
\end{abstract}

\section{Introduction}
\footnotetext[1]{These authors contributed equally to this work.}
\footnotetext[2]{Corresponding author.} 
Large language models (LLMs) \cite{achiam2023gpt, chowdhery2023palm, touvron2023llama1} have undergone impressive advancements that transform NLP tasks into a unified text-to-text paradigm, achieving robust alignment and generation capabilities through reinforcement learning from human feedback (RLHF) \cite{ouyang2022training, bai2022training}.
Particularly, models are required to predict the correct labels in natural language understanding (NLU) tasks, distinct from natural language generation (NLG) tasks. 
Numerous studies have employed ``rationales'' to assist LLMs with Chain-of-Thought (CoT) prompting during supervised fine-tuning (SFT) stage \cite{kim2023cot, hsieh2023distilling}. Rationale refers to the relevant parts or information that provide explanations or support for the predictions or decisions made by a model. 

\begin{figure}
    \centering
    \includegraphics[width=1\linewidth]{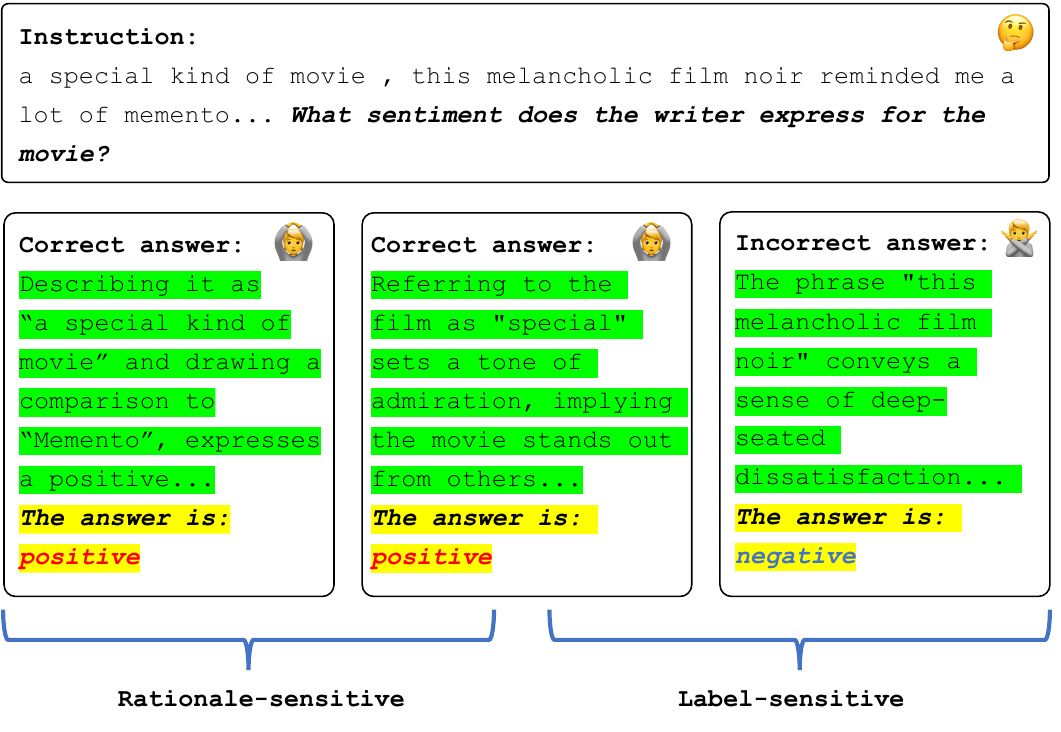}
    \caption{The example of rationale-sensitive and label-sensitive pairs from sentiment classification. Highlight rationales in green and labels in yellow.}
    \label{fig:pair_example}
\end{figure}

\begin{figure}
    \centering
    \includegraphics[width=1\linewidth]{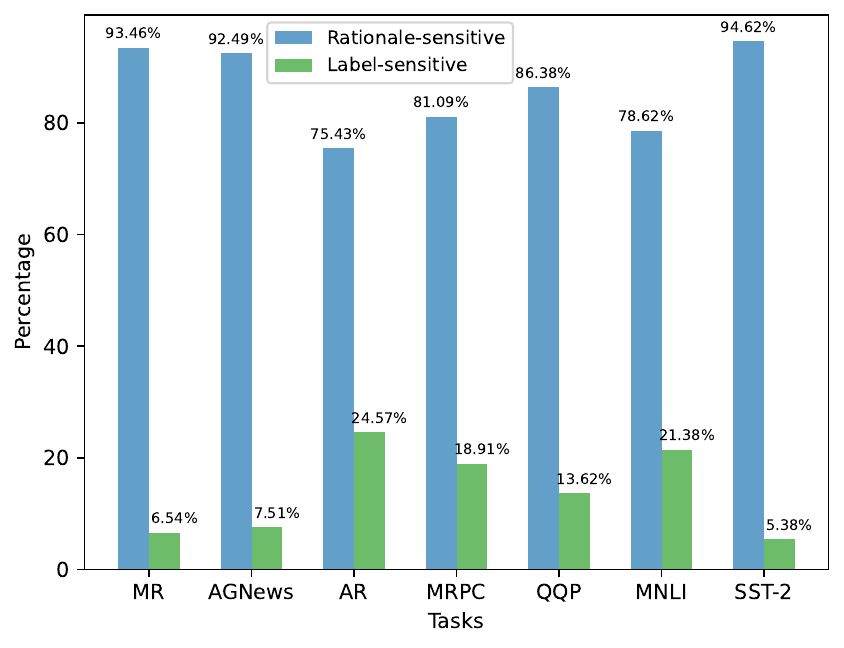}
    \caption{The distribution of rationale-sensitive and label-sensitive pairs sampled from SFT model across a range of tasks.}
    \label{fig:rlhf_pair_distribution}
\end{figure}

However, \citet{lambert2023alignment} detail a fundamental challenge in RLHF learning schemes: the \textit{objective mismatch} issue. This arises when the reward model is influenced by human preference data, introducing biases that conflict with downstream evaluation metrics, especially when applied to NLU tasks. 
In RLHF, comparison data is initially sampled from the SFT model and ranked by a labeler. Then the policy model is optimized against the reward model that is trained with these pairs to align with human preference.
For NLU tasks, the pairs can be categorized into rationale-sensitive and label-sensitive. As illustrated in Figure \ref{fig:pair_example}, we provide an example where three answers sampled from the SFT model for the same instruction. If two answers have the same label and different rationales, they form a rationale-sensitive pair, with the more reasonable rationale considered superior. In contrast, if two answers have different labels, they form a label-sensitive pair, with the correct label deemed superior. 
However, we observed that the pairs sampled from the SFT model mainly fall into the category of rationale-sensitive. 
Figure \ref{fig:rlhf_pair_distribution} shows the specific distribution ratios of pairs across several NLU tasks. The percentage of rationale-sensitive pairs exceeds 75\%, and in datasets like SST-2, MR, and AGNews, surpasses 90\%. 
The severe imbalance in the distribution of pairs leads the model to prioritize the quality of rationales over the correctness of labels during RLHF training, which conflicts with the evaluation metric (mostly \textit{label accuracy}) of NLU tasks. A detailed analysis is presented in Section \ref{sec:main-results}.

To address this challenge, our paper proposes a Reinforcement Learning framework enhanced with Label-sensitive Reward (RLLR) for NLU tasks. Firstly, we leverage GPT-4 to generate rationales corresponding to the gold labels of the training data. The SFT model is trained with rationales, incorporating CoT prompting to enhance comprehension abilities. Secondly, we generate rationales for the incorrect labels (relative to the gold labels). Unlike RLHF, which uses human intervention to rank sentences, RLLR automatically constructs label-sensitive pairs for training the reward model based on the correctness of the label. The comparison data is initially sampled from the trained SFT model. Finally, we train the policy model against the label-sensitive reward model with Proximal Policy Optimization (PPO) to prioritize the correctness of labels. 
Furthermore, optimizing with mixed rewards from the label-sensitive and rationale-sensitive reward models, RLLR$_\textsc{mixed}$ ensures both the accuracy of labels and the quality of rationales.
Extensive experiments on eight NLU tasks demonstrate that our method consistently outperforms the SFT baseline by an average of 1.54\% and the RLHF baseline by an average of 0.69\%, while also exhibiting higher quality in rationales generation.

Our contributions are summarized as:

(1) We propose a Reinforcement Learning framework enhanced with Label-sensitive Reward (RLLR) for NLU tasks to tackle the \textit{objective mismatch} issue.

(2) Optimizing with mixed rewards,  RLLR$_\textsc{mixed}$ can achieve promising performance on both the accuracy of labels and the quality of rationales. 

(3) Through empirical experiments, we demonstrate the effectiveness of our method.  We have conducted a thorough investigation into various aspects, including the utilization of rationales, the performance of reward models, the quality of generated rationales and a detailed case study.

\begin{table*}[!ht]
  \centering
  \resizebox{1.0\linewidth}{!}{
    \begin{tabular}{p{1.0\linewidth}}
    \toprule
    \textbf{Original example} \\
    \midrule
    Wall St. Bears Claw Back Into the Black (Reuters) Reuters - Short-sellers, Wall Street's dwindling band of ultra-cynics, are seeing green again. \\
    \textbf{Correct label}: \textbf{\textcolor[RGB]{0,0,255}{Business}}; \textbf{Incorrect labels}: \textbf{\textcolor[RGB]{255,0,0}{World Politics, Sports, Science and Technology}} \\
    \midrule
    \textbf{Prompt for generating rationales for correct label} \\
    \midrule
    What label best describes this news article? \\
    Wall St. Bears Claw Back Into the Black (Reuters) Reuters - Short-sellers, Wall Street's dwindling band of ultra-cynics, are seeing green again. \\
    Please give a rationale for the answer \textbf{\textcolor[RGB]{0,0,255}{``Business''}} in a confident tone (regardless of the true answer): \\
    \midrule
    \textbf{Prompt for generating rationales for incorrect label} \\
    \midrule
    What label best describes this news article? \\
    Wall St. Bears Claw Back Into the Black (Reuters) Reuters - Short-sellers, Wall Street's dwindling band of ultra-cynics, are seeing green again. \\
    Please give a rationale for the answer \textbf{\textcolor[RGB]{255,0,0}{``World Politics''}} in a confident tone (regardless of the true answer): \\
    \bottomrule
    \end{tabular}
  }
  \caption{Demonstration of label-sensitive pair generation process on AGNews. First, we generate a rationale for the correct label ``Business''. Then we randomly select an incorrect label ``World Politics'', and generate a rationale for it. The table shows the prompts for requesting GPT-4 to generate corresponding rationales. While we use GPT-4 for rationale generation, this approach is adaptable to manual annotation or alternative methods.}
  \label{tab:pair_costruction_demo}
\end{table*}

\section{Related Work}

\paragraph{Reinforcement Learning from Human Feedback.} LLMs have demonstrated commendable performance, leveraging RLHF to achieve notable alignment and generation capabilities \cite{ouyang2022training, achiam2023gpt, bai2022training, ziegler2019fine}. 
RLHF aims to optimize the policy language model to generate content that is desired by humans. 
Recently, some research endeavors have uncovered inherent challenges in RLHF \cite{casper2023open,lambert2023history}, including feedback type limitations, evaluation difficulties, oversight challenges, etc. 
Several methods have been proposed to mitigate these challenges.
\citet{bai2022constitutional} introduce RL from AI Feedback (RLAIF), training an AI assistant through self-improvement while adhering to constitutional principles that constrain model-generated content. 
\citet{wu2023fine} introduce a fine-grained RLHF framework that uses fine-grained human feedback, such as identifying false sentences or irrelevant sub-sentences, as an explicit training signal.
\citet{rafailov2024direct} introduced a new parameterization of the reward model in RLHF that enables extraction of the corresponding optimal policy in closed form. This allows us to solve the standard RLHF problem with only a simple classification loss.
\citet{song2024preference} proposed Preference Ranking Optimization (PRO) as an alternative to PPO for directly aligning LLMs with the Bradley-Terry comparison to accommodate preference rankings of any length.
However, these approaches encounter a fundamental challenge in RLHF learning schemes: the \textit{objective mismatch} issue \cite{lambert2023alignment}. 
In this paper, we tackle this problem by training the reward model with the label-sensitive pairs.

\paragraph{Chain-of-Thought.} CoT can significantly improve the complex reasoning ability of LLMs by generating natural language rationales that lead to the final answer \cite{wei2022chain, kim2023cot}. 
\citet{hsieh2023distilling} introduce a distilling mechanism step-by-step,  extracting LLM rationales as additional supervision for training small models within a multi-task framework.
\citet{fu2023specializing} propose a method to specialize the model’s ability (smaller than 10B) towards a target task with CoT prompting. In this paper, we enhance the performance of LLMs on NLU tasks with CoT prompting utilizing rationales generated for the labels.

\begin{figure*}[ht]
    \centering
    \includegraphics[width=1\linewidth]{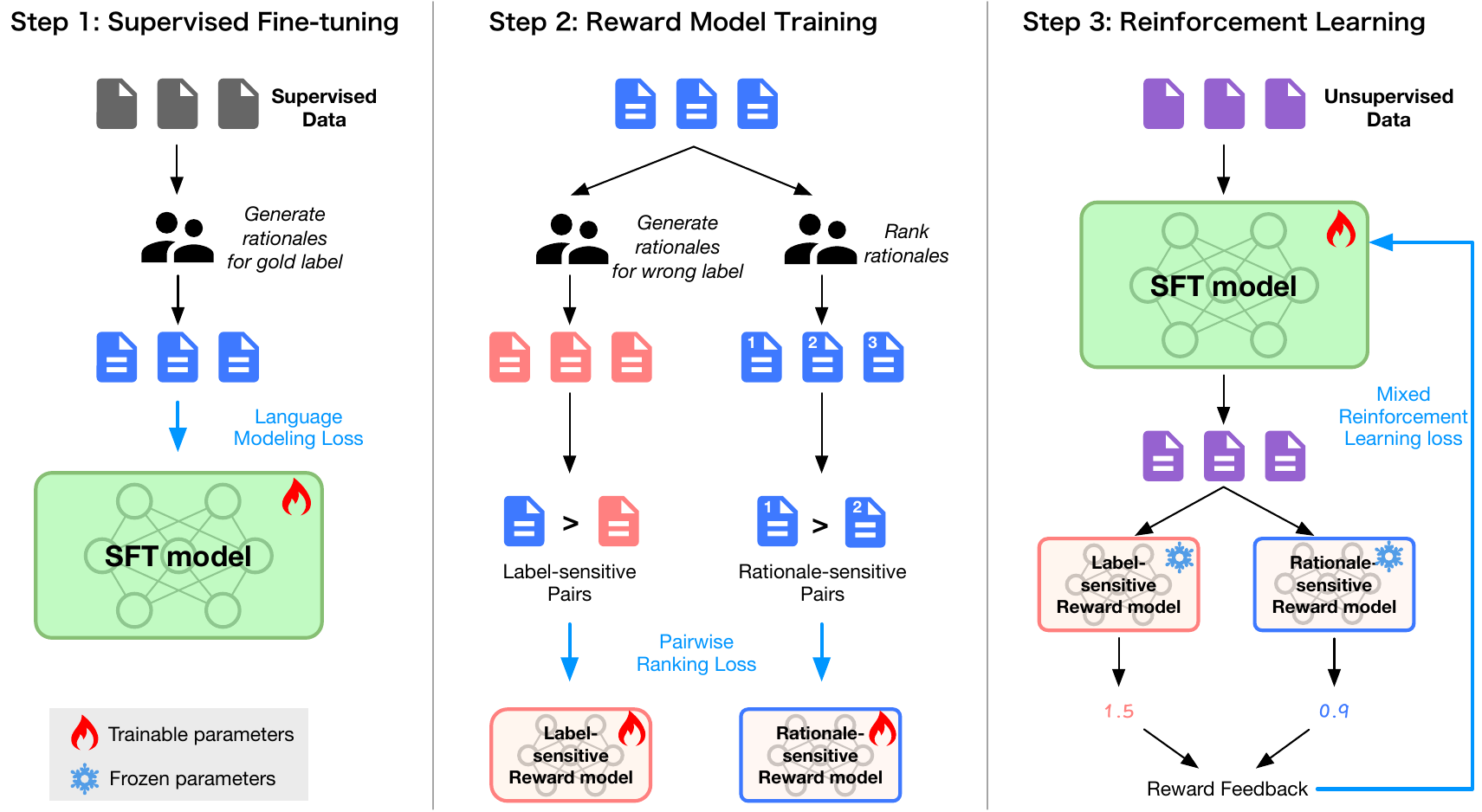}
    \caption{The training pipeline of RLLR with supervised fine-tuning, reward model training, and mixed reinforcement learning. Blue arrows indicate data used for model training.}
    \label{fig:model}
\end{figure*}

\section{Proposed Method}

In this section, we introduce the training pipeline of our method as illustrated in Figure \ref{fig:model}, including supervised fine-tuning, reward model training, and reinforcement learning enhanced with mixed rewards.
\subsection{Supervised Fine-Tuning}
In NLU tasks, the supervised dataset is denoted as $\mathcal{S}=\{x,y\}$, where $x$ denotes the sentence and $y$ denotes the class label. The unsupervised dataset is denoted as $\mathcal{U}$, and the foundation model is denoted as $\pi$.
According to \citet{wei2022chain}, generating rationales that lead to the final answer can significantly improve the reasoning ability of LLMs through CoT. Therefore, we first generate a rationale from the sentence $x$ and the label $y$ with a specific prompt template using either human annotators or LLMs such as GPT-4. Then we reform the original dataset $\mathcal{S}$ to the training dataset $\mathcal{T}=\{q, a\}$. The question $q$ is constructed by $x$ with a template and the answer $a$ with $t$ tokens is obtained by combining rationale and label, denoted as $a = a_{1, \cdots, t}$. The details of prompts can be found in Appendix \ref{sec:appendix_prompts_of_tasks} and \ref{sec:appendix_rationale_collection}.
The foundation model $\pi$ is then trained on $\mathcal{T}$ to obtain the model $\pi_\text{SFT}$. Formally, the loss for supervised fine-tuning is defined as:
\begin{equation}
\mathcal{L}_{\text{SFT}} = -\mathbb{E}_{(q, a) \sim \mathcal{T}}\left[\log P_{\pi} \left(a_t \mid q, a_{1, \cdots, t-1}\right)\right].
\end{equation}

\subsection{Reward Model Training}
In the second phase, comparison data are sampled from the answers generated by the SFT model $\pi_\text{SFT}$ given a question. 
As illustrated in Figure \ref{fig:rlhf_pair_distribution}, more than 75\% of the pairs generated by the SFT model are rationale-sensitive pairs (i.e., both answers have the same label). 
The sentences in the rationale-sensitive pair are then labeled with a preference order.
In RL, the reward model denoted as $r_\phi$ assigns higher scores to preferable answers compared with unfavorable ones, employing the Bradley-Terry paired comparison \cite{bradley1952rank}.
In this scenario, the model prioritizes the quality of the generated rationales over the accuracy of the labels. This focus shift results in less than optimal performance, stemming from the previously mentioned issue of \textit{objective mismatch}.

To address this issue, we generate rationales based on the incorrect label for an input sentence and combine them to form a new answer. We leverage GPT-4 to generate rationales for the correct label and incorrect label.
we generate rationales for incorrect labels $\hat{y}$ to create a new answer $\hat{a}$, which is a rationale-augmented incorrect answer. Along with the correct answer $a$, we can obtain the preferences $a \succ \hat{a} \mid q$ for label-sensitive pairs without extra annotation. The process of the rationale-sensitive pair generation is illustrated in Table \ref{tab:pair_costruction_demo}, and additional details can be found in Appendix \ref{sec:appendix_label_sensitive_pair}. The label-sensitive and rationale-sensitive pairs are used to train two reward models, respectively. Specifically, we have $a^1 \succ a^2 \mid q$ to represent the preference in the pair. To predict these preferences, we employ the Bradley-Terry (BT) model, which defines the preference probability as follows: 
\begin{equation}
    P_{BT}=\frac{\exp\left(r_{\phi}\left(q, a^1\right)\right)}{\exp\left (r_{\phi}\left(q, a^1\right)\right)+\exp\left (r_{\phi}\left(q, a^2\right)\right)}.
\end{equation}
This objective is framed as a binary classification problem to train the reward model $r_\phi(q, a)$ with the loss defined as:
\begin{equation}
\resizebox{1\linewidth}{!}{
$\mathcal{L}_R = -\mathbb{E}_{\left(q, a^1, a^2\right) \sim \mathcal{C}}\left[\log \sigma\left(r_\phi\left(q, a^1\right) - r_\phi\left(q, a^2\right)\right)\right]$,}
\end{equation}
where $\sigma$ is the logistic function and $\mathcal{C}$ is the dataset of comparisons. In this way, we can obtain two separate reward models $r_{\phi1}$ and $r_{\phi2}$ with the label-sensitive and rationale-sensitive pairs, respectively. The reward model $r_\phi(q, a)$ is often initialized from the SFT model $\pi_\text{SFT}(a|q)$ with the addition of a linear layer on top of the final transformer layer that produces a single scalar prediction for the reward value.

\subsection{Reinforcement Learning}
During the RL phase, we use the reward model to train the SFT model $\pi_\text{SFT}$ using Proximal Policy Optimization (PPO) on the unsupervised dataset $\mathcal{U}$. Given a question constructed by the sentence from $\mathcal{U}$, the mixed reward function 
from $r_{\phi1}(q, a)$ and $r_{\phi2}(q, a)$ is calculated as :
\begin{equation}
\resizebox{1\linewidth}{!}{
$r_{\textsc{m}}(q, a) =
\begin{cases} 
  r_{\phi1}(q, a) + r_{\phi2}(q, a), & \text{if $r_{\phi1}(q, a) < \lambda$}\\  
  \lambda +  r_{\phi2}(q, a), & \text{if $r_{\phi1}(q, a) \ge \lambda$}
\end{cases}$
}
\end{equation}
where $\lambda$ is a hyper-parameter as the threshold for $r_{\phi1}$ (the label-sensitive reward model). According to experimental observations, the reward score of $r_{\phi1}$ converges to around 5.0, while the score of $r_{\phi2}$ is within 1.0, resulting in an imbalance between the two. To prevent reinforcement learning from being completely dominated by $r_{\phi1}$, we set a threshold value $\lambda$. 
When the score of $r_{\phi1}$ is less than $\lambda$, the combined reward score is the sum of $r_{\phi1}$ and $r_{\phi2}$; when the score of $r_{\phi1}$ is greater than or equal to $\lambda$, the combined reward score is equal to $\lambda$ plus $r_{\phi2}$. 
We first optimize the policy based on $r_{\phi1}$, focusing on the correctness of the labels. As the RL training progresses, the score of $r_{\phi1}$ gradually exceeds $\lambda$. Once the score of $r_{\phi1}$ surpasses $\lambda$, we truncate it. At this point, the model will pay more attention to $r_{\phi2}$, which is the quality of the rationale.
In this way, both $r_{\phi1}$ and $r_{\phi2}$ can play a role in reinforcement learning, allowing the final policy model to predict the correct labels while generating high-quality rationales. 

To guide the RL training, the loss function is constructed by combining the rewards generated by the reward model with a KL divergence constraint, which ensures that the policy does not deviate significantly from its initial behavior, defined as:
\begin{equation}
\resizebox{1\linewidth}{!}{
$
\max\limits_{\pi_\mathrm{RL}}\mathbb{E}_{(q, a) \sim D_{\pi_{\mathrm{RL}}}}\left[r_\textsc{m}(q, a)- \beta \log \left(\frac{\pi_{\mathrm{RL}}(a \mid q)}{\pi_{\mathrm{SFT}}(a \mid q)}\right)\right]$,
}
\end{equation}
where $\pi_\mathrm{RL}$ is the learned RL policy, $\pi_{\mathrm{SFT}}$ is the SFT model, and $\beta$ is the KL reward efficient controlling the strength of the KL penalty. RLLR$_\textsc{mixed}$ is obtained from this objective with two reward models while RLLR is trained only with the label-sensitive rationale reward model. 

\section{Experiments}

\subsection{Experiment Setup}


\paragraph{Datasets.} We evaluate the performance of our proposed method across eight NLU tasks, encompassing five from the GLUE benchmark \cite{wang2018glue}. The tasks include Movie Reviews (MR) \cite{pang-lee-2005-seeing}, AppReviews (AR) \cite{Zurich2017Review} and SST-2 for sentiment classification, AGNews \cite{zhang2015character} for topic classification, MRPC and QQP for paraphrase detection, MNLI for textual entailment, and STS-B for semantic similarity. We employ the Pearson correlation coefficient as our evaluation metric for STS-B, and accuracy for others. To ensure a fair comparison with baseline methods, we convert all tasks into a text-to-text format following \cite{sanh2021multitask}. For methods that require rationales, we utilize GPT-4 to generate rationales conditioned on given labels.

To ensure our model is free from cognitive biases, we refine our process with GPT-4 through meticulous manual reviews and prompt template adjustments. We manually evaluate the annotation through sampling and designed specific strategies to filter out data that did not meet our standards. Several specific challenges have been encountered, such as GPT-4's reluctance to justify certain labels, particularly incorrect ones, and discrepancies between its generated rationales and the assigned labels. We address these issues by iteratively refining the prompt template and developing a classifier to filter out corrupt data. The details regarding the prompt templates and rationale annotation process are provided in Appendix \ref{sec:appendix_prompts_of_tasks} and \ref{sec:appendix_rationale_collection}.

\begin{table*}[!ht]
  \centering
  \resizebox{1\linewidth}{!}{
    \begin{tabular}{llrrrrrrrrr}
    \toprule
    Splits / Tasks & Unit & MR & AGNews & AR & MRPC & QQP & MNLI & SST-2 & STS-B & ALL \\
    \midrule
    SFT Train & Prompts & 2,000 & 5,000 & 5,000 & 1,000 & 5,000 & 5,000 & 5,000 & 5,000 & 38,000 \\
    RLHF-RM Train & Pairs & 16,740 & 15,993 & 12,620 & 9,683 & 16,956 & 15,757 & 16,475 & 15,751 & 119,975 \\
    RLHF-PPO Train & Prompts & 5,000 & 5,000 & 4,426 & 2,668 & 5,000 & 5,000 & 5,000 & 1,323 & 33,417 \\
    RLLR-RM Train & Pairs & 12,339 & 12,349 & 12,339 & 8,877 & 12,318 & 12,323 & 12,338 & 12,317 & 95,200 \\
    RLLR-PPO Train & Prompts & 5,000 & 5,000 & 4,426 & 2,668 & 5,000 & 5,000 & 5,000 & 1,323 & 33,417 \\
    Test & Prompts & 1,000 & 1,000 & 1,000 & 408 & 1,000 & 2,000 & 872 & 1,000 & 8,280 \\
    \bottomrule
    \end{tabular}
  }
  \caption{The number of examples used in experiments.}
  \label{tab:num_of_examples}
\end{table*}

\paragraph{Baselines.} We conduct our experiments using several state-of-the-art foundation models, including LLaMA2 \cite{touvron2023llama}, Baichuan2 \cite{yang2023baichuan}, ChatGLM3 \cite{du2021glm}, Mistral \cite{jiang2023mistral}, and Bloom \cite{workshop2022bloom}. We compare our method with two prevalent training methods: (1) SFT, which refines LLMs through optimization against a conditional language modeling objective on supervised data; (2) RLHF, which involves training a reward model on preference data and subsequently employing this model to guide RL-based fine-tuning. For RLHF, we utilize GPT-4 for the preference annotation within our experiments. Detailed procedural information can be found in Appendix \ref{sec:appendix_preference_collection}.

\paragraph{Training.} To streamline the experimental complexity, we fine-tune the models on a multi-task dataset, rather than on datasets for individual tasks. To address the task imbalance issue, we construct the training set at a maximum of 5,000 samples per task. The surplus examples are used as unsupervised data for PPO in RLHF and RLLR. This approach mirrors real-world scenarios where unsupervised data is abundant, but supervised data is scarce. We also construct a multi-task test set comprising up to 1,000 examples from each task to enhance experimental efficiency without compromising validity. The details of the examples are listed in Table \ref{tab:num_of_examples}. In all experiments, we employ Low-Rank Adaptation (LoRA) \cite{hu2021lora} fine-tuning, as opposed to full-parameter tuning, achieving up to an 80\% reduction in GPU memory requirements. Within the RL-based approaches, the policy, reward, and value models are equipped with their own set of LoRA parameters.

We employ 4 V100 GPUs, each with 32 GB of memory, to train our 7B models. To maintain a consistent batch size across various experiments, we utilize the gradient accumulation. The Adam optimizer, coupled with a cosine schedule, is used in all experiments, with a fixed LoRA rank of 16. We perform a hyperparameter search within a limited range and observe that the performance exhibits minimal sensitivity to changes in these hyperparameters. The ranges for the primary hyperparameters, such as learning rate, batch size, and training epochs, are detailed in Table \ref{tab:hyperparam}.

\begin{table}[!ht]
  \centering
  \resizebox{1\linewidth}{!}{
    \begin{tabular}{lccc}
    \toprule
    Stage & Learning Rate & Batch Size & Epochs \\
    \midrule
    SFT & 1e-5\textasciitilde2e-5 & 128 & 20 \\
    SFT w. rat. & 1e-4\textasciitilde1e-3 & 128 & 10 \\
    RLHF Reward & 2e-4 & 64\textasciitilde128 & 10 \\
    RLHF PPO & 2e-6\textasciitilde1e-5 & 16\textasciitilde32 & 1 \\
    RLLR Reward & 1e-4\textasciitilde1e-3 & 64\textasciitilde128 & 1 \\
    RLLR PPO & 2e-6\textasciitilde1e-5 & 16\textasciitilde32 & 1 \\
    \bottomrule
    \end{tabular}
  }
  \caption{Range of hyperparameters.}
  \label{tab:hyperparam}
\end{table}


\begin{table*}[!ht]
  \centering
  \resizebox{1\linewidth}{!}{
    \begin{tabular}{llccccccccc}
    \toprule
    \multicolumn{2}{c}{Methods / Dataset} & MR & AGNews & AR & MRPC & QQP & MNLI(m/mm) & SST-2 & STS-B & AVG. \\
    \midrule
    \multirow{5}{*}{LLaMA2 7B} & SFT & 91.00 & 92.20 & 69.40 & 82.11 & 85.50 & 83.50/\underline{85.10} & 96.22 & 89.24 & 86.03 \\
       & SFT \textit{w. rat.} & 91.90 & 92.50 & 68.70 & 83.58 & 87.90 & 83.50/85.00 & \underline{96.56} & 91.83 & 86.74 \\
       & RLHF & 91.90 & 93.00 & 68.50 & \underline{83.82} & 87.60 & \underline{83.60}/85.00 & 96.44 & 92.02 & 86.79 \\
       \cmidrule{2-11}
       & RLLR & \underline{92.40} & \underline{93.40} & \textbf{70.10} & \underline{83.82} & \textbf{88.20} & \textbf{85.10}/\textbf{85.90} & \textbf{96.79} & \textbf{92.31} & \textbf{87.47} \\
       & RLLR$_\textsc{mixed}$ & \textbf{92.60} & \textbf{93.50} & \underline{69.60} & \textbf{84.07} & \underline{88.00} & \textbf{85.10}/\textbf{85.90} & \textbf{96.79} & \underline{92.07} & \underline{87.40} \\
    \midrule
    \multirow{5}{*}{ChatGLM3 6B} & SFT & 89.00 & 93.00 & 68.80 & 81.37 & 85.00 & 81.80/83.90 & \underline{95.30} & 89.79 & 85.33 \\
       & SFT \textit{w. rat.} & \underline{91.30} & \underline{93.10} & 68.20 & 81.86 & 84.90 & 82.80/84.20 & \textbf{95.87} & 90.10 & 85.51 \\
       & RLHF & 91.10 & \underline{93.10} & \underline{68.90} & \underline{82.35} & 85.00 & 82.80/\underline{84.30} & \textbf{95.87} & 90.14 & 85.64 \\
       \cmidrule{2-11}
       & RLLR & \textbf{91.40} & \textbf{93.40} & \textbf{69.10} & \underline{82.35} & \underline{85.50} & \textbf{83.60}/\textbf{84.60} & \textbf{95.87} & \textbf{91.12} & \textbf{86.02} \\
       & RLLR$_\textsc{mixed}$ & \textbf{91.40} & \textbf{93.40} & \textbf{69.10} & \textbf{82.36} & \textbf{85.70} & \underline{83.50}/\textbf{84.60} & \textbf{95.87} & \underline{90.91} & \underline{85.69} \\
    \midrule
    \multirow{5}{*}{Mistral 7B} & SFT & 92.10 & 92.50 & \underline{70.40} & 83.58 & 85.90 & 84.70/87.50 & 95.18 & 91.17 & 87.00 \\
       & SFT \textit{w. rat.} & 92.00 & \underline{92.70} & 69.40 & 86.52 & 86.10 & 85.40/87.60 & 96.33 & 92.06 & 87.29 \\
       & RLHF & 92.10 & 92.20 & 68.70 & 85.29 & \underline{88.30} & 85.40/87.80 & 96.22 & 91.83 & 87.26 \\
       \cmidrule{2-11}
       & RLLR & \textbf{93.30} & \textbf{93.10} & \textbf{70.60} & \textbf{87.01} & \underline{88.30} & \underline{86.60}/\textbf{88.90} & \textbf{96.90} & \textbf{92.32} & \textbf{88.27} \\
       & RLLR$_\textsc{mixed}$ & \underline{92.40} & \underline{92.70} & 70.30 & \underline{86.76} & \textbf{88.70} & \textbf{86.80}/\underline{88.80} & \underline{96.67} & \underline{92.23} & \underline{88.10} \\
    \midrule
    \multirow{5}{*}{Baichuan2 7B} & SFT & 90.80 & \textbf{93.40} & 69.90 & 81.86 & 84.90 & 82.90/84.10 & 95.64 & 89.09 & 85.84 \\
       & SFT \textit{w. rat.} & 90.70 & 92.50 & 69.10 & 82.35 & 87.00 & 84.80/85.00 & 95.99 & 91.58 & 86.19 \\
       & RLHF & \underline{91.20} & 92.90 & 68.30 & \underline{83.09} & 86.50 & 84.50/85.30 & 96.22 & 91.50 & 86.25 \\
       \cmidrule{2-11}
       & RLLR & \textbf{91.30} & \underline{93.00} & \underline{70.40} & 82.84 & \underline{87.40} & \textbf{85.70}/\textbf{85.80} & \underline{96.33} & \textbf{91.94} & \underline{86.82} \\
       & RLLR$_\textsc{mixed}$ & \underline{91.20} & \underline{93.00} & \textbf{70.50} & \textbf{83.58} & \textbf{87.50} & \underline{85.50}/\underline{85.70} & \textbf{96.44} & \underline{91.81} & \textbf{86.88} \\
    \midrule
    \multirow{5}{*}{Bloom 7B} & SFT & 89.20 & 89.80 & 69.30 & 76.96 & 83.40 & 75.80/78.50 & 94.38 & 87.88 & 82.80 \\
       & SFT \textit{w. rat.} & 89.50 & 91.80 & 69.80 & 82.60 & 83.60 & 76.50/\underline{80.70} & \underline{94.61} & 88.58 & 83.92 \\
       & RLHF & \underline{89.70} & \textbf{92.70} & 69.40 & 82.11 & \underline{84.00} & 77.00/80.00 & \underline{94.61} & \underline{88.96} & 84.01 \\
       \cmidrule{2-11}
       & RLLR & \textbf{90.10} & \textbf{92.70} & \textbf{70.90} & \underline{84.07} & \textbf{84.30} & \textbf{77.90}/\textbf{81.30} & \textbf{95.53} & \textbf{89.04} & \textbf{84.83} \\
       & RLLR$_\textsc{mixed}$ & 89.50 & \underline{92.50} & \underline{70.40} & \textbf{84.31} & \textbf{84.30} & \underline{77.80}/\underline{80.70} & \underline{94.61} & 88.95 & \underline{84.52} \\
    \bottomrule
    \end{tabular}
  }
  \caption{Experiment results for our methods and baselines, over a range of foundation models and NLU tasks. The abbreviation ``SFT \textit{w. rat.}'' stands for SFT with rationale.}
  \label{tab:main}
\end{table*}


\subsection{Main Results}
\label{sec:main-results}

Our main experiment results are shown in Table \ref{tab:main}. Additional results for models of various sizes are available in Appendix \ref{sec:appendix_different_size}. SFT \textit{w. rat.} and SFT denote models fine-tuned on supervised data with and without rationales, respectively. RLHF denotes models fine-tuned with the standard RLHF procedure, which predominantly utilizes rationale-sensitive pairs. RLLR denotes models fine-tuned using our proposed method, with a reward model trained on label-sensitive pairs. RLLR$_\textsc{mixed}$ further integrates reward models trained on both label-sensitive and rationale-sensitive pairs. 
The policy model is initialized from the SFT \textit{w. rat.} model in both RLHF, RLLR, and RLLR$_\textsc{mixed}$ settings.

Comprehensive evaluations across five foundational models and eight NLU tasks reveal that our RLLR method consistently surpasses the SFT baseline by an average margin of 1.54\%, and the RLHF baseline by an average of 0.69\%. The maximum average improvement over RLHF was achieved on Mistral 7B, reaching 1.02\%. The enhancement observed in ChatGLM3 6B, while modest, is still quantifiable at an increase of 0.38\%. RLLR and RLLR$_\textsc{mixed}$ also achieve the best results on most individual tasks, except Baichuan2 on AGNews. However, integrating RLLR with other models consistently yields a performance enhancement on AGNews, most notably, exceeding the SFT baseline by 2.9\% with Bloom-7B. This substantial improvement robustly validates the efficacy of the proposed method.

The integration of rationales brings improvement over the vanilla SFT by an average margin of 0.79\%, demonstrating the benefit of rationales. Despite this improvement, the performance of SFT \textit{w. rat.} still lags behind that of RLLR, suggesting that simply integrating the SFT method with rationales is insufficient. Moreover, the RLHF baseline mirrors the performance of SFT \textit{w. rat.}, with no additional gains, which corroborates the presence of an \textit{objective mismatch} issue.

In Section \ref{sec:analysis}, we further analyze the influence of various mechanisms, including the utilization of rationales, reward modeling objectives, and incorporation of multiple rewards in RL fine-tuning. The RLLR$_\textsc{mixed}$ method achieves on-par performance with RLLR, surpassing SFT by an average of 1.45\%, and RLHF by an average margin of 0.60\%. However, we further examine its impact on the quality of rationales, extending our analysis beyond label accuracy.

\subsection{Analysis}
\label{sec:analysis}

\paragraph{Utilization of rationales.} Incorporating rationales into the SFT stage achieves improvement across 78\% of our task-model pairings (35 out of 45), aligning with the advancements reported by \cite{hsieh2023distilling, kim2023cot, fu2023specializing}. Nonetheless, a comparative analysis between SFT with RLLR indicates that the mere addition of rationales to SFT is insufficient. SFT, categorized under Behavior Cloning within the Imitation Learning framework, is prone to suffering from compounding errors \cite{ross2011reduction}. Theoretically, the minimum expected error for a policy derived through Behavior Cloning grows quadratically with the length of the trajectories. Introducing rationales under this method paradoxically extends trajectory lengths, exacerbating the issue. In contrast, RLLR, rooted in Inverse Reinforcement Learning, effectively reduces compounding errors by optimizing across entire trajectories rather than individual actions \cite{ho2016generative, swamy2023inverse}, thereby enhancing the effectiveness of rationales.

\begin{figure*}[!ht]
\centering
\subfigure[RLHF vs. SFT]{\includegraphics[width=0.45\linewidth]{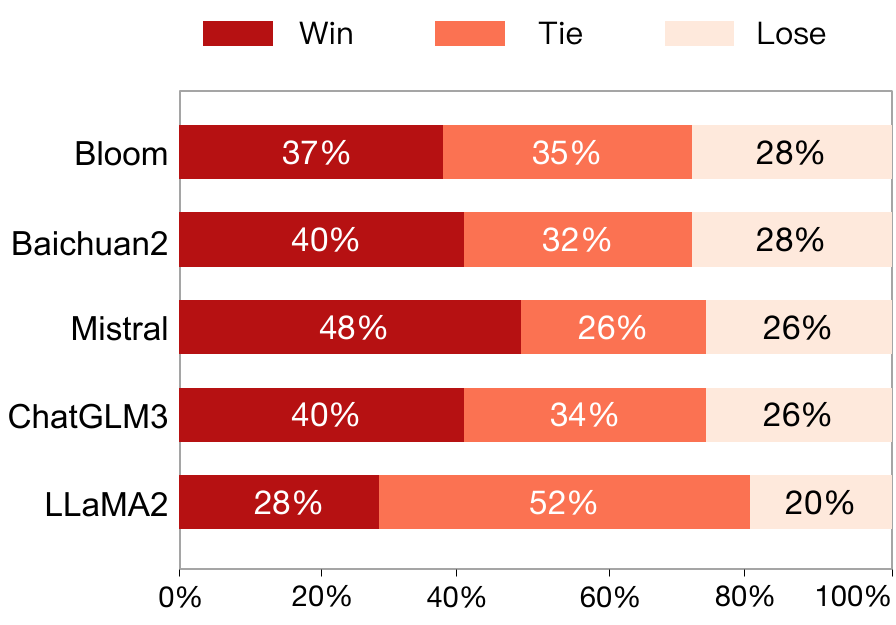}} \hspace{0.5cm}
\subfigure[RLLR vs. SFT]{\includegraphics[width=0.45\linewidth]{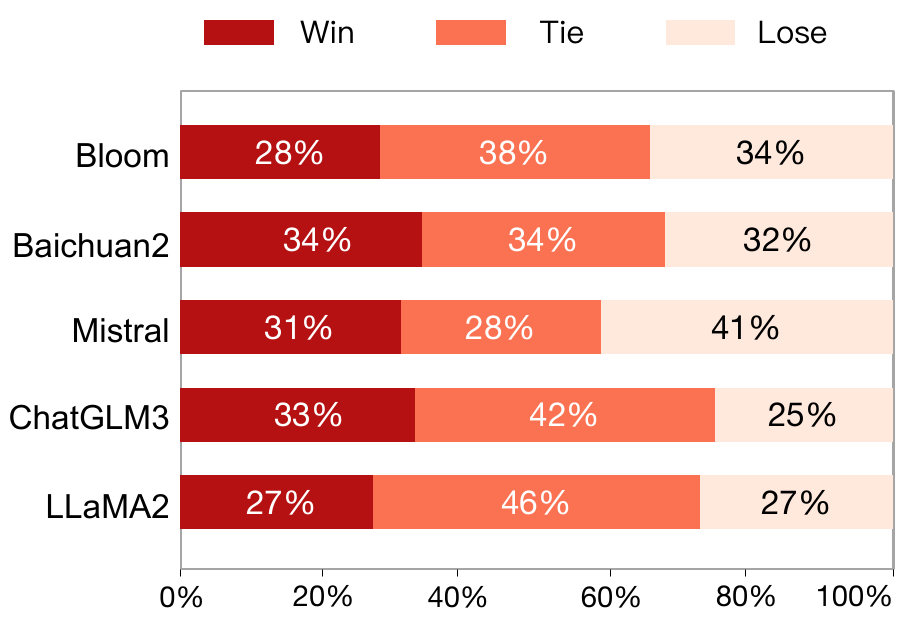}}
\subfigure[RLLR$_\textsc{mixed}$ vs. SFT]{\includegraphics[width=0.45\linewidth]{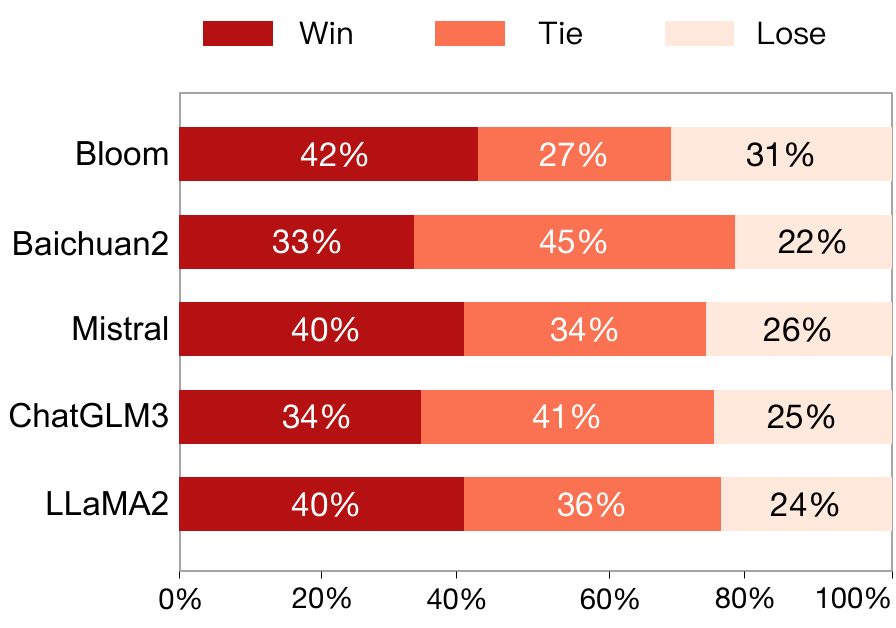}}\hspace{0.5cm}
\subfigure[RLLR$_\textsc{mixed}$ vs. RLHF]{\includegraphics[width=0.45\linewidth]{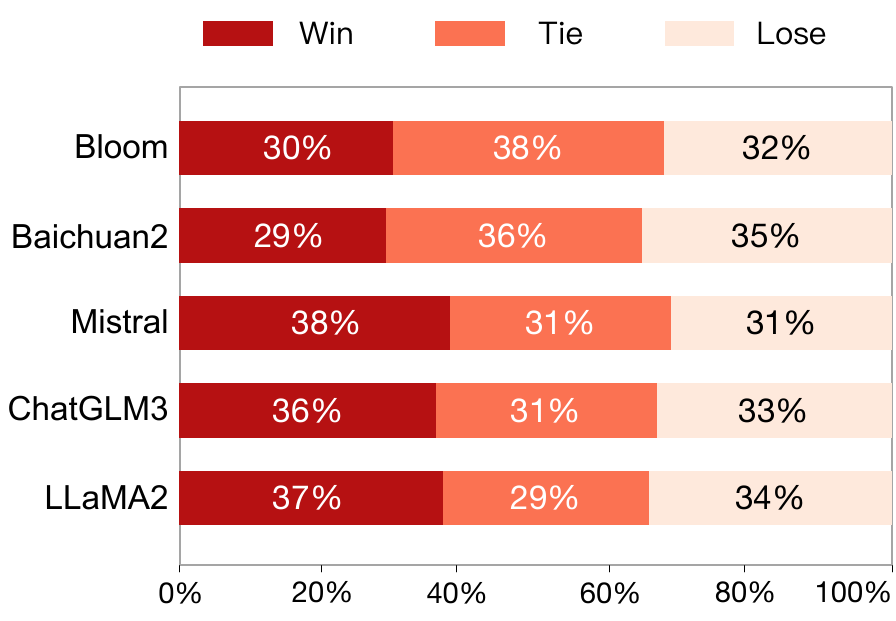}}
\caption{Evaluation of rationale quality judged by GPT-4, compared to the SFT and RLHF methods.}\label{fig:win_rate}
\end{figure*}

\paragraph{Reward model performance.} To elucidate the superiority of RLLR over RLHF, we scrutinized the efficacy of reward models trained in both methods. The models are evaluated on a hold-out label-sensitive dataset, comprising pairs of correct and incorrect answers with respect to the gold label. This evaluation framework is designed to assess the models' proficiency in differentiating between rationales that lead to either the correct or incorrect labels. As indicated in Table \ref{tab:reward_performance}, reward models developed under RLLR demonstrate an average accuracy of 90\%, outperforming those from RLHF by a margin of 10\%, which stand at an average accuracy of 80\%. Detailed results on each individual task are presented in Appendix \ref{sec:appendix_reward_performance}. These findings align with the main results and underscore the detrimental impact of \textit{objective mismatch} issue within RLHF. Conversely, RLLR is immune to such discrepancies, as its objectives are congruent with the evaluative criteria used to discern between correct and incorrect rationales, thereby yielding superior outcomes across a spectrum of tasks and models.

\begin{table}[htbp]
    \centering
    \resizebox{1\linewidth}{!}{
      \begin{tabular}{lcc}
      \toprule
      Models / Training Set & RLHF reward & RLLR reward \\
      \midrule
      LLaMA2 7B & 80.92 & 91.66 \\
      ChatGLM3 6B & 75.00 & 90.20 \\
      Mistral 7B & 80.78 & 91.39 \\
      Baichuan2 7B & 81.46 & 90.18 \\
      Bloom 7B & 77.75 & 88.73 \\
      \bottomrule
      \end{tabular}%
    }
    \caption{Performance of reward models on hold-out label-sensitive pairs.}
    \label{tab:reward_performance}%
\end{table}%


\begin{table*}[!ht]
  \centering
  \resizebox{1\linewidth}{!}{
    \begin{tabular}{p{0.1\linewidth}p{0.9\linewidth}}
    \toprule
    Question & Movie review: ... mafia , rap stars and hood rats butt their ugly heads in a regurgitation of cinematic violence that gives brutal birth to an unlikely , but likable , hero . \textbf{\textit{What sentiment dose the movie review express?}} \\
    \midrule
    RLHF & The movie review expresses a decidedly negative sentiment. The language used is laced with derogatory terms and a tone of disdain, which clearly indicates the reviewer's unfavorable opinion of the film... \textbf{\textcolor[RGB]{0,0,255}{Furthermore, the use of ``likable, hero'' in a seemingly sarcastic or begrudging manner does not redeem the overall negative impression}}... \textbf{\textit{The answer is: negative}} \\
    RLLR & The movie review expresses a positive sentiment, \textbf{\textcolor[RGB]{255,0,0}{as it highlights an ``unlikely, but likable, hero'' that emerges from the chaotic and violent setting}}... The use of the word ``likable'' is a clear indicator of the reviewer's approval, suggesting that the character's journey and the film's outcome are enjoyable and endearing... \textbf{\textit{The answer is: positive}} \\
    \bottomrule
    \toprule
    Question & \textbf{\textit{Analyze the semantic similarity between two sentences and predict a similarity score from 0 to 5.}} Sentence 1: Someone is greating a carrot., Sentence 2: A woman is grating an orange food. \\
    \midrule
    RLLR & \textbf{\textcolor[RGB]{0,0,255}{Firstly, the subjects}} of the sentences are different... \textbf{\textcolor[RGB]{0,0,255}{Secondly, the actions}} described in the sentences are not the same... \textbf{\textcolor[RGB]{0,0,255}{Thirdly, the objects}} of the sentences are distinct... \textbf{\textcolor[RGB]{0,0,255}{Lastly, the contexts}} implied by the sentences are different... Given these points of divergence, a similarity score of 1.000 accurately captures the minimal semantic overlap between the two sentences... \textbf{\textit{The answer is: 1.000}} \\
    RLLR$_\textsc{mixed}$ & \textbf{\textcolor[RGB]{255,0,0}{1. **Subjects**}}: In the first sentence... \textbf{\textcolor[RGB]{255,0,0}{2. **Actions**}}: The actions described in both sentences are related but not identical... \textbf{\textcolor[RGB]{255,0,0}{3. **Objects**}}: The objects of the actions are different... \textbf{\textcolor[RGB]{255,0,0}{4. **Context**}}: The context of the actions is also different... \textbf{\textcolor[RGB]{255,0,0}{5. **Semantic Field**}}: Both sentences fall within the semantic field of food-related activities, but... Given these points, a score of 1.000 accurately captures the low level of semantic similarity between the two sentences... \textbf{\textit{The answer is: 1.000}} \\
    \bottomrule
    \end{tabular}
  }
  \caption{Examples generated by RLHF, RLLR, and RLLR$_\textsc{mixed}$ methods.}
  \label{tab:case_study}
\end{table*}

\paragraph{Quality of generated rationales.}
\label{para:quality}

Despite the modest enhancement in accuracy, RLHF has significantly advanced the quality of text generation by integrating human preferences during fine-tuning. Beyond the accuracy on NLU tasks, the quality of the generated text is also a key consideration. This is particularly relevant in human-in-the-loop contexts, where the model's output serves as a guide for human operators, necessitating text that is both high-quality and reflective of human values. To assess this quality, we examined a subset of queries from the validation set and appraised the response quality produced by various models with GPT-4. We employed the win rate against SFT \textit{w. rat.} as a metric for evaluation. The evaluation results are shown in Figure \ref{fig:win_rate} and the implementation details are described in Appendix \ref{sec:appendix_eval_of_quality}. The RLHF method outperforms SFT in generating high-quality rationales, evidenced by a 39\% win rate, a 26\% lose rate, and a 36\% tie rate on average. In contrast, the RLLR method demonstrates comparable performance to SFT, with a balanced win or lose ratio of 30\% on average. This outcome aligns with expectations, as RLLR did not incorporate any human preference data. The hybrid RLLR$_\textsc{mixed}$ method, which integrates the reward models of both RLHF and RLLR, achieves a 38\% win rate, a 26\% lose rate, and a 37\% tie rate on average against the SFT baseline. When compared to RLHF, RLLR$_\textsc{mixed}$ also exhibits a comparable efficacy, with a win rate of 34\% and a lose rate of 33\%. Notably, RLLR$_\textsc{mixed}$ yields the most favorable outcomes, delivering rationale quality equivalent to RLHF and maintaining label accuracy at the level of RLLR.


\paragraph{Case study.}
\label{para:case_study}

Table \ref{tab:case_study} showcases examples of answers generated by the RLHF, RLLR, and RLLR$_\textsc{mixed}$ methods. For the sake of brevity, we have omitted any superfluous content from the answers, retaining only the essential context and comparative elements. To visually highlight the primary differences within the examples, we've used blue and red color coding.
In the first example, which is sampled from the Movie Review dataset, the reviewer first describes the negative aspects of the movie, but at the end, a turning point is made by proposing ``give birth to an unlikely but likable hero'' to express positive sentiments. The RLHF model fails to recognize a positive sentiment shift, despite the reviewer's concluding praise, leading to an incorrect negative assessment. 
On the other hand, the RLLR model accurately identifies the shift in sentiment, leading to an accurate evaluation. The second example from the STS-B dataset demonstrates that while the RLLR and RLLR$_\textsc{mixed}$ methods yield similar similarity scores for identical sentence pairs, the RLLR$_\textsc{mixed}$ approach enhances the rationale's comprehensiveness by integrating an extra``Semantic Field'' component. Furthermore, the RLLR$_\textsc{mixed}$ output utilizes \textbf{Markdown} formatting to enhance readability. These findings suggest that the RLLR$_\textsc{mixed}$ method can significantly improve the rationale's quality compared to the standard RLLR approach, which proves our viewpoint.

\section{Conclusion}
In this paper, we introduce a Reinforcement Learning framework
enhanced with Label-sensitive Reward to amplify the performance of LLMs for NLU. 
By training the reward model on label-sensitive pairs, which are constructed by generating rationales for the incorrect labels, we mitigate the \textit{objective mismatch} issue in RLHF, leading to improved performance in NLU tasks. 
Extensive results on 5 foundation models and 8 NLU tasks demonstrate that RLLR consistently surpasses the SFT baseline by a margin of 1.54\%, and the RLHF baseline by 0.69\%. 
By additionally incorporating the label-sensitive and rationale-sensitive rewards, our enhanced RLLR$_\textsc{mixed}$ method not only maintains the label accuracy comparable to RLLR but also achieves rationale quality on par with RLHF. 
We present an in-depth analysis of RLLR, examining the utilization of rationales, reward modeling objectives, and incorporation of multiple rewards during the RL stage. 
The results and analysis substantiate the effectiveness of our methods in the NLU tasks.

\section{Limitations}

Due to cost considerations, there are some deficiencies in our work, which we have listed here for future reference. Firstly, integrating rationales into model responses increases computing power requirements and generation time as a trade-off for enhanced accuracy and interpretability. Secondly, we utilize GPT-4 as a proxy of humans to generate rationales, annotate preferences, and evaluate the quality of rationales. Despite the success made by advanced AI models like GPT-4 in supplanting manual annotation, we believe that experiments with authentic human annotation and evaluation remain essential. Finally, the compatibility of RLLR with RL-free methods such as DPO, PRO, and RRHF remains unexplored. We leave these limitations for future work.



\bibliography{anthology,custom}
\clearpage

\appendix

\begin{table*}[!ht]
  \centering
  \resizebox{1.0\linewidth}{!}{
    \begin{tabular}{p{0.1\linewidth}p{0.9\linewidth}}
      \toprule
      Question & yet another weepy southern bore-athon . \textbf{\textit{What sentiment does the writer express for the movie?}} \\
      \midrule
      Label & \textbf{\textit{negative}} \\
      \midrule
      Rationale & The sentiment expressed by the writer for the movie is decidedly negative. The term ``weepy'' suggests that the movie is overly sentimental or melodramatic in a way that the writer finds unappealing. Additionally, ``bore-athon'' is a pejorative play on the word ``marathon'' indicating that the writer perceives the movie as boring and tedious, and possibly that it drags on for too long. The use of the word ``another'' implies a sense of frustration or fatigue with what the writer views as a recurrent theme or genre, in this case, a ``southern'' one, which the writer apparently does not find engaging or innovative. The overall tone is dismissive and conveys a strong negative opinion about the film. \\
      \bottomrule
      \toprule
      Question & \textbf{\textit{What label best describes this news article? \textbackslash n}} LABOUR AT BRIGHTON: SORRY I JUST CAN \#39;T SAY SORRY FOR IRAQ TONY Blair refused to say sorry for the Iraq war yesterday, after a last-minute change of mind over a conference apology. In his Brighton address, the Prime Minister \\
      \midrule
      Label & \textbf{\textit{World Politics}} \\
      \midrule
      Rationale & The label ``World politics'' aptly describes the news article as it directly involves a statement by Tony Blair, who was the Prime Minister of the United Kingdom, concerning a matter of international significance—the Iraq war. The reference to Blair's address at a conference in Brighton and his stance on a major foreign policy issue that has had global repercussions places the article firmly within the realm of international relations and governmental affairs. The content is inherently political, dealing with decisions and actions that have shaped world events, and it pertains to the political landscape beyond just a national level, thereby making ``World politics'' the most fitting label for this news piece. \\
      \bottomrule
      \toprule
      Question & \textbf{\textit{Does the sentence \textbackslash n}} As they wound through police barricades to the funeral home , many chanted ``Celia, Celia'' and sang snippets of her songs . \textbf{\textit{\textbackslash n paraphrase (that is, mean the same thing as) this sentence? \textbackslash n}} As they wound through police barricades to the funeral home , many chanted ``Celia, Celia'' \\
      \midrule
      Label & \textbf{\textit{no}} \\
      \midrule
      Rationale & No, the two sentences do not convey the exact same meaning. While both sentences describe a procession moving through police barricades to a funeral home with people chanting ``Celia, Celia'' the first sentence includes an additional action that the second sentence omits. Specifically, the first sentence mentions that many in the crowd also sang snippets of her songs. This additional detail about singing introduces a separate action that captures more about the atmosphere and the tribute being paid to Celia. It is not merely a repetition of her name but an engagement with her musical legacy. Without this element, the second sentence fails to capture the full extent of the homage being paid, thus it is not a complete paraphrase of the first. \\
      \bottomrule
    \end{tabular}
  }
  \caption{Examples of generated rationales.}
  \label{tab:rationale_example}
\end{table*}

\section{Prompts for Tasks}
\label{sec:appendix_prompts_of_tasks}

For all of the tasks, we use the following template for SFT w. rationale, RLHF and RLLR:

\noindent \texttt{\{\{rationale\}\}\textbackslash n\textbackslash n The answer is: \{\{label\}\}}

\noindent The name in double curly brackets represents a variable and should be replaced with its value. Input templates and possible labels for each task are listed below.

\subsection{Movie Reviews}

Input template:

\noindent \texttt{\{\{text\}\} What sentiment does the writer express for the movie?}

\noindent Possible labels:

\noindent \texttt{negative, positive}

\subsection{AGNews}

Input template:

\noindent \texttt{What label best describes this news article? \textbackslash n \{\{text\}\}}

\noindent Possible labels:

\noindent \texttt{World politics, Sports, Business, Science and technology}

\subsection{MNLI}

Input template:

\noindent \texttt{Given a premise and a hypothesis, predict the relationship between them. Choose one of the following labels: entailment, contradiction, or neutral. Premise:\{\{sentence1\}\}, Hypothesis:\{\{sentence2\}\}}

\noindent Possible labels:

\noindent \texttt{entailment, contradiction, neutral}

\subsection{QQP}

Input template:

\noindent \texttt{I received the questions ``\{\{sentence1\}\}'' and ``\{\{sentence2\}\}''. Are they duplicates?}

\noindent Possible labels:

\noindent \texttt{no, yes}

\subsection{SST-2}

Input template:

\noindent \texttt{Movie review: \{\{text\}\} What sentiment dose the movie review express?}

\noindent Possible labels:

\noindent \texttt{negative, positive}

\subsection{STS-B}

Input template:

\noindent \texttt{Analyze the semantic similarity between two sentences and predict a similarity score from 0 to 5. Sentence 1: \{\{sentence1\}\}, Sentence 2: \{\{sentence2\}\}}

\noindent Possible labels:

\noindent Float number in range [0.0, 5.0].

\subsection{MRPC}

Input template:

\noindent \texttt{Does the sentence \textbackslash n \{\{sentence1\}\} \textbackslash n paraphrase (that is, mean the same thing as) this sentence? \textbackslash n \{\{sentence2\}\}}

\noindent Possible labels:

\noindent \texttt{no, yes}

\subsection{AppReviews}

Input template:

\noindent \texttt{On a scale of 1-5 (with 1 being least favorable and 5 being most favorable), how would you rate this review? ``\{\{text\}\}''}

\noindent Possible labels:

\noindent \texttt{1, 2, 3, 4, 5}

\section{Rationale Collection}
\label{sec:appendix_rationale_collection}

We utilize the following prompt templates to request GPT-4 for rationales. Sometimes GPT-4 refuses to give a rationale conditioned the provided label, and we train a simple classifier to filter out these responses. Examples of generated rationales are shown in Table \ref{tab:rationale_example}.

\noindent \texttt{\{\{question\}\} \textbackslash n\textbackslash n Please give a rationale for the answer ``\{\{label\}\}'' in a confident tone (regardless of the true answer):}

\section{Preference Collection}
\label{sec:appendix_preference_collection}

We sample 5 responses from the SFT model for each example and ask GPT-4 to rank the responses. The prompt template for requesting GPT-4 is as follows:

\noindent \texttt{Given the following question and answers, please rank the answers according to your preference, considering accuracy, coherence, logicality, factuality, relevance, and information completeness. \textbackslash n\textbackslash n [Question] \{question\} \textbackslash n\textbackslash n [Answer 1] \{answer 1\} \textbackslash n\textbackslash n [Answer 2] \{answer 2\} \textbackslash n\textbackslash n [Answer 3] \{answer 3\} \textbackslash n\textbackslash n [Answer 4] \{answer 4\} \textbackslash n\textbackslash n [Answer 5] \{answer  5\} \textbackslash n\textbackslash n Please give your rationale first, and then give the ranking. Output format: ``\{rationale\} \textbackslash n\textbackslash n Ranking: \{e.g. \{\{ranking\_example\}\}\}''}

\noindent The variable \texttt{\{\{ranking\_example\}\}} is generated by shuffling the list \texttt{[1, 2, 3, 4, 5]} and concatenating them with ``>'' or ``='', e.g. \texttt{5>3>2>4>1} or \texttt{2>1=5>3=4}. We generate a different example for every GPT-4 request to avoid bias.

\section{Construction of Label-Sensitive Pairs}
\label{sec:appendix_label_sensitive_pair}

In tasks involving categorical labels, an incorrect label is randomly chosen from the full label set excluding the correct label, to create a label-sensitive pair. For the AppReviews and STS-B tasks, which use a rating scale from 0 to 5, incorrect labels are generated by adding 3 to the correct label and then incorporating a random value from the range [-1, 1]. For instance, given a correct STS-B label of 2.8, a random increment of 0.3 is selected, resulting in an initial incorrect label of 2.8+3+0.3=6.1. This exceeds the maximum rating, so we adjust by subtracting 5, yielding a final incorrect label of 1.1. In the case of the AppReviews task, this label is subsequently rounded to an integer.

\begin{table*}[!htbp]
  \centering
  \resizebox{1.0\linewidth}{!}{
    \begin{tabular}{lllccccccccc}
    \toprule
    \multicolumn{3}{c}{Methods / Dataset} & MR & AGNews & AR & MRPC & QQP & MNLI(m/mm) & SST-2 & STS-B & AVG. \\
    \midrule
    \multirow{10}{*}{LLaMA2} & \multirow{5}{*}{7B} & SFT & 91.00 & 92.20 & 69.40 & 82.11 & 85.50 & 83.50/\underline{85.10} & 96.22 & 89.24 & 86.03 \\
       &    & SFT \textit{w. rat.} & 91.90 & 92.50 & 68.70 & 83.58 & 87.90 & 83.50/85.00 & \underline{96.56} & 91.83 & 86.74 \\
       &    & RLHF & 91.90 & 93.00 & 68.50 & \underline{83.82} & 87.60 & \underline{83.60}/85.00 & 96.44 & 92.02 & 86.79 \\
       \cmidrule{3-12}
       &    & RLLR & \underline{92.40} & \underline{93.40} & \textbf{70.10} & \underline{83.82} & \textbf{88.20} & \textbf{85.10}/\textbf{85.90} & \textbf{96.79} & \textbf{92.31} & \textbf{87.47} \\
       &    & RLLR$_\textsc{mixed}$ & \textbf{92.60} & \textbf{93.50} & \underline{69.60} & \textbf{84.07} & \underline{88.00} & \textbf{85.10}/\textbf{85.90} & \textbf{96.79} & \underline{92.07} & \underline{87.40} \\
    \cmidrule{2-12}
       & \multirow{5}{*}{13B} & SFT & 92.00 & 92.20 & 69.00 & 81.62 & 88.00 & 83.10/85.20 & 96.33 & 90.12 & 86.40 \\
       &    & SFT \textit{w. rat.} & 92.20 & 92.40 & 68.90 & 83.82 & 88.40 & 85.40/87.10 & \textbf{96.79} & \underline{91.78} & 87.42 \\
       &    & RLHF & 92.20 & 92.70 & 68.90 & \textbf{85.78} & \underline{88.70} & \underline{85.60}/87.20 & \underline{96.67} & 91.49 & 87.69 \\
       \cmidrule{3-12}
       &    & RLLR & \textbf{92.60} & \textbf{93.00} & \underline{69.50} & \underline{85.54} & \textbf{88.80} & \textbf{86.10}/\textbf{87.80} & \textbf{96.79} & \textbf{92.23} & \textbf{88.04} \\
       &    & RLLR$_\textsc{mixed}$ & \underline{92.40} & \underline{92.90} & \textbf{69.70} & \underline{85.54} & \textbf{88.80} & \textbf{86.10}/\underline{87.50} & \textbf{96.79} & \textbf{92.23} & \underline{88.00} \\
    \midrule
    \multirow{10}{*}{Bloom} & \multirow{5}{*}{3B} & SFT & 88.40 & 90.20 & 67.90 & 75.25 & 81.30 & 73.30/74.70 & \underline{93.46} & 86.66 & 81.24 \\
       &    & SFT \textit{w. rat.} & 88.70 & \underline{92.00} & 68.50 & \underline{80.15} & 81.90 & 73.40/75.10 & 93.23 & 86.58 & 82.17 \\
       &    & RLHF & 88.60 & \underline{92.00} & 68.60 & 79.66 & 82.20 & 74.20/75.60 & 93.00 & 86.23 & 82.23 \\
       \cmidrule{3-12}
       &    & RLLR & \textbf{89.80} & \textbf{92.30} & \underline{69.20} & \textbf{80.64} & \textbf{82.60} & \underline{74.60}/\textbf{76.70} & \underline{93.46} & \textbf{87.58} & \textbf{82.99} \\
       &    & RLLR$_\textsc{mixed}$ & \underline{89.40} & \underline{92.00} & \textbf{69.30} & \textbf{80.64} & \underline{82.40} & \textbf{74.70}/\underline{76.40} & \textbf{93.58} & \underline{87.17} & \underline{82.84} \\
    \cmidrule{2-12}
       & \multirow{5}{*}{7B} & SFT & 89.20 & 89.80 & 69.30 & 76.96 & 83.40 & 75.80/78.50 & 94.38 & 87.88 & 82.80 \\
       &    & SFT \textit{w. rat.} & 89.50 & 91.80 & 69.80 & 82.60 & 83.60 & 76.50/\underline{80.70} & \underline{94.61} & 88.58 & 83.92 \\
       &    & RLHF & \underline{89.70} & \textbf{92.70} & 69.40 & 82.11 & \underline{84.00} & 77.00/80.00 & \underline{94.61} & \underline{88.96} & 84.01 \\
       \cmidrule{3-12}
       &    & RLLR & \textbf{90.10} & \textbf{92.70} & \textbf{70.90} & \underline{84.07} & \textbf{84.30} & \textbf{77.90}/\textbf{81.30} & \textbf{95.53} & \textbf{89.04} & \textbf{84.83} \\
       &    & RLLR$_\textsc{mixed}$ & 89.50 & \underline{92.50} & \underline{70.40} & \textbf{84.31} & \textbf{84.30} & \underline{77.80}/\underline{80.70} & \underline{94.61} & 88.95 & \underline{84.52} \\
    \bottomrule
    \end{tabular}
  }
  \caption{Results of varying sized models.}
  \label{tab:model_size_analysis}
\end{table*}
\section{Results of Varying Sized Models}
\label{sec:appendix_different_size}

To substantiate the scalability of our method across models of varying sizes, we also conduct a series of experiments using LLaMA2-13B and Bloom-3B models. The results are presented in Table \ref{tab:model_size_analysis}, in conjunction with those of the 7B models to facilitate direct comparison. For LLaMA, the 7B model's performance improved by 0.68\% and the 13B model improved by 0.35\% over the RLHF baseline. For BLOOM, the 3B model's performance improved by 0.76\% and the 7B model improved by 0.82\%. Interestingly, smaller models don't always get greater improvements. This consistency across disparate model sizes strongly supports the scalability of our proposed RLLR method.

\section{Reward Model Performance}
\label{sec:appendix_reward_performance}

Table \ref{tab:reward_performance_full} presents the performance of reward models trained with RLHF and RLLR on eight individual NLU tasks. Reward models employing RLLR methods demonstrated an overall accuracy of approximately 90\%, surpassing those trained with RLHF by a significant margin of 10 percentage points, with the latter achieving an accuracy of 80\%. The gap between RLLR and RLHF on STS-B and AppReviews tasks is most significant, exceeding 35\% and 20\% respectively. The gap on AGNews, MRPC, and QQP tasks also exceeds 5\%, indicating that RLHF suffers from \textit{objective mismatch} issue on these tasks.

\section{Evaluation of Generation Quality}
\label{sec:appendix_eval_of_quality}
\begin{table*}[ht]
  \centering
  \resizebox{1.0\linewidth}{!}{
    \begin{tabular}{lcccccccccc}
    \toprule
    \multirow{2}{*}{Tasks / Models} & \multicolumn{2}{c}{LLaMA2 7B} & \multicolumn{2}{c}{ChatGLM3 6B} & \multicolumn{2}{c}{Mistral 7B} & \multicolumn{2}{c}{Baichuan2 7B} & \multicolumn{2}{c}{Bloom 7B} \\
       & RLHF & RLLR & RLHF & RLLR & RLHF & RLLR & RLHF & RLLR & RLHF & RLLR \\
    \midrule
    MR & 90.13 & 92.07 & 80.26 & 90.94 & 90.94 & 90.78 & 89.32 & 91.59 & 91.10 & 88.51 \\
    AGNews & 89.67 & 96.31 & 85.61 & 94.10 & 87.45 & 96.68 & 89.67 & 94.46 & 88.56 & 94.10 \\
    AR & 69.91 & 92.76 & 74.91 & 91.26 & 68.91 & 92.51 & 67.79 & 90.26 & 65.29 & 90.89 \\
    MRPC & 84.03 & 86.58 & 71.25 & 87.22 & 76.04 & 86.58 & 77.96 & 83.07 & 78.91 & 82.43 \\
    QQP & 79.82 & 86.63 & 70.18 & 86.63 & 78.92 & 87.53 & 80.46 & 88.17 & 76.74 & 83.80 \\
    MNLI & 86.65 & 90.75 & 83.00 & 89.04 & 87.40 & 91.13 & 88.59 & 89.56 & 84.41 & 87.47 \\
    SST-2 & 92.24 & 94.68 & 82.93 & 93.13 & 93.13 & 94.01 & 93.13 & 92.90 & 89.36 & 92.24 \\
    STS-B & 58.68 & 97.07 & 46.80 & 93.05 & 62.34 & 94.52 & 64.90 & 92.50 & 50.46 & 93.97 \\
    \midrule
    Overall & 80.92 & 91.66 & 75.00 & 90.20 & 80.78 & 91.39 & 81.46 & 90.18 & 77.75 & 88.73 \\
    \bottomrule
    \end{tabular}
  }
  \caption{Performance of reward models on hold-out label-sensitive pairs. Results across five different foundation models are presented.}
  \label{tab:reward_performance_full}
\end{table*}
We utilize GPT-4 as a proxy for human evaluation. First, we sample a set of questions and corresponding answers generated by two methods. To mitigate positional bias, we then randomize the order of the answers within each pair. The question along with two answers is subsequently formatted according to the predefined GPT4 input template:
\noindent \texttt{Given the following question and two candidate answers, please choose which one is better, considering accuracy, coherence, logicality, factuality, relevance, and information completeness. \textbackslash n \textbackslash n [Question] \{question\} \textbackslash n \textbackslash n [Answer 1] \{answer 1\} \textbackslash n \textbackslash n [Answer 2] \{answer 2\} \textbackslash n \textbackslash n Please response with ``Answer 1 is better'' or ``Answer 2 is better'' or ``Equal'' first, and then give your rationale.}

\end{document}